\documentclass[11pt, a4paper, copyright, logo, nonumbering]{ipads}
\usepackage[bottom]{footmisc}
\usepackage[authoryear, sort&compress, round]{natbib}
\usepackage{dblfloatfix}
\usepackage{caption}
\usepackage{xspace}
\usepackage{pifont}
\usepackage{multirow}
\usepackage{tcolorbox}
\usepackage{xltabular}
\usepackage{longtable}
\usepackage{hyperref}
\usepackage{makecell}
\usepackage{scalefnt}
\usepackage{amsfonts}
\usepackage{amsmath}
\usepackage{amssymb}
\usepackage{lineno}
\usepackage{multirow}
\usepackage{adjustbox}

\usepackage{subfig}
\usepackage[bottom]{footmisc}

\usepackage{CJKutf8}
\usepackage{setspace}

\usepackage{dsfont}
\usepackage{array} 
\usepackage{tabularx} 
\usepackage{xcolor} 

\usepackage{hyperref}
\hypersetup{
    colorlinks,
    linkcolor={black},
    citecolor={blue!50!black},
    urlcolor={blue!80!black}
}

\usepackage{lipsum}  
\usepackage{multicol} 

\makeatletter
\def\@BTrule[#1]{%
  \ifx\longtable\undefined
    \let\@BTswitch\@BTnormal
  \else\ifx\hline\LT@hline
    \nobreak
    \let\@BTswitch\@BLTrule
  \else
     \let\@BTswitch\@BTnormal
  \fi\fi
  \global\@thisrulewidth=#1\relax
  \ifnum\@thisruleclass=\tw@\vskip\@aboverulesep\else
  \ifnum\@lastruleclass=\z@\vskip\@aboverulesep\else
  \ifnum\@lastruleclass=\@ne\vskip\doublerulesep\fi\fi\fi
  \@BTswitch}
\makeatother

\addto\extrasenglish{
}

 {\begin{list}{}%
         {\setlength{\leftmargin}{#1}}%
         \item[]%
 }
 {\end{list}}

\reportnumber{001} 

\renewcommand{\today}{}
\newcommand{\model}{SmallThinker}

\newcommand\rurl[1]{%
  \href{http://#1}{\nolinkurl{#1}}%
}

\title{\centering {\model: A Family of Efficient Large Language Models Natively Trained for Local Deployment}}

\author[*]{
\vspace{-2em}~\\
\small\texttt{Institute of Parallel and Distributed Systems, Shanghai Jiao Tong University}\\
\small\texttt{School of Artificial Intelligence, Shanghai Jiao Tong University}\\
\small\texttt{Zenergize AI}
}

\renewcommand{\phi}{\varphi}

\renewcommand{\epsilon}{\varepsilon}
\renewcommand{\imath}{\mathrm{i}}

\newlength{\restsubwidth}
\newlength{\restsubheight}
\newlength{\restsubmoreheight}
\newcommand{\rest}[2]{%
        \settowidth{\restsubwidth}{\ensuremath{#2}}
        \settoheight{\restsubheight}{\ensuremath{{}_{#2}}}
        \ensuremath{{#1\hskip 0.5pt}_{\vrule\kern2pt\parbox[b][%
        4pt][b]{\the\restsubwidth}{%
                        \ensuremath{{}_{#2}}}}}
        }

\begin{document}
\begin{CJK*}{UTF8}{gbsn}

\begin{abstract}

\paragraph{Abstract}
While frontier large language models (LLMs) continue to push capability boundaries, 
  their deployment remains confined to GPU-powered cloud infrastructure.
  We challenge this paradigm with \model{}, a family of LLMs natively designed—not adapted—for 
  the unique constraints of local devices: weak computational power, limited memory, and slow storage.
  Unlike traditional approaches that mainly compress existing models built for clouds, 
  we architect \model{} from the ground up to thrive within these limitations.
  Our innovation lies in a deployment-aware architecture that transforms constraints into design principles.
  First, We introduce a two-level sparse structure combining fine-grained Mixture-of-Experts (MoE) with sparse feed-forward networks,
  drastically reducing computational demands without sacrificing model capacity.
  Second, to conquer the I/O bottleneck of slow storage, we design a pre-attention router that enables 
  our co-designed inference engine to prefetch expert parameters from storage while computing attention, effectively 
  hiding storage latency that would otherwise cripple on-device inference.
  Third, for memory efficiency, we utilize NoPE-RoPE hybrid sparse attention mechanism to slash KV cache requirements.
We release \model-4B-A0.6B and \model-21B-A3B, which achieve state-of-the-art performance scores and even outperform larger LLMs.
  Remarkably, our co-designed system mostly eliminates the need for expensive GPU hardware: 
  with Q4\_0 quantization, both models exceed 20 tokens/s on ordinary consumer CPUs, while consuming 
  only 1GB and 8GB of memory respectively.
  \model{} is publicly available at \rurl{hf.co/PowerInfer/SmallThinker-4BA0.6B-Instruct} and \rurl{hf.co/PowerInfer/SmallThinker-21BA3B-Instruct}.
\end{abstract}

\maketitle

\begin{figure}[h]
    \centering
    \includegraphics[width=1\linewidth]{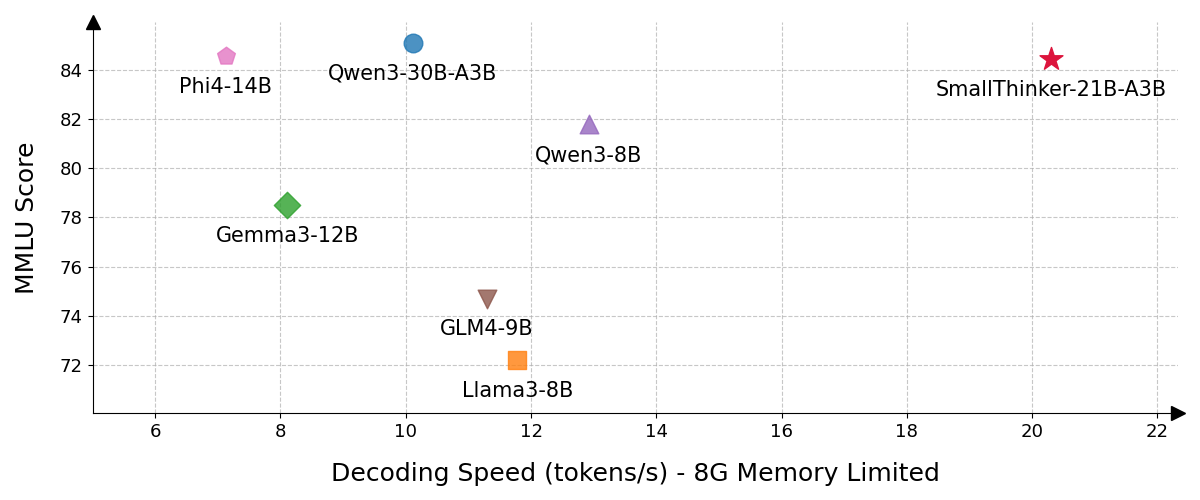}
    \caption{Comparative analysis of inference performance and MMLU scores. \model{} achieves SOTA performance, outperforming comparable models in both speed and accuracy.}
    \label{fig:perf_vs_speed}
\end{figure}

\newpage

\begin{spacing}{0.9}
\tableofcontents
\end{spacing}

\newpage

\section{Introduction}

The AI revolution has reached an inflection point.
While frontier models like GPT-4, Gemini~\citep{comanici2025gemini},
and Claude demonstrate unprecedented capabilities,
they remain prisoners of the cloud—tethered to massive data centers with unlimited power and memory.
Yet the future of AI lies not only in distant servers,
but also in the devices we carry every day: smartphones processing sensitive conversations privately,
laptops running complex analyses offline,
and local devices bringing intelligence to billions without internet access.
This vision demands a fundamental rethinking of how we build large language models.

The prevailing paradigm for on-device AI involves adapting models originally architected for large-scale, high-memory GPU clusters. This post-hoc adaptation, typically via methods like distillation, or pruning, inherently trades model capability for on-device feasibility. This creates a dichotomy between powerful, cloud-based models and their significantly compromised on-device counterparts.
We reject this problematic choice. Instead of asking ``how can we squeeze cloud models onto local devices?'',
we ask a more fundamental question: ``what would an large language model (LLM) look like if designed from first principles for local constraints?''

This question led us to train \model{} from ground up,
a family of LLMs natively architected for local deployment.
Rather than treating weak computational power, limited memory, and slow storage as obstacles to overcome,
we embrace them as design principles that shape every architectural decision.
As shown in Figure~\ref{fig:perf_vs_speed}, the result is not a compromised cloud model, but a new class of AI system that thrives within local constraints while delivering frontier-level capabilities.

Our key insight is that local constraints, when properly understood, reveal opportunities for novel architectural innovations.
The limited compute budget motivated our two-level sparse structure that combining fine-grained Mixture-of-Experts (MoE)
with sparse feed-forward networks and only activating essential parameters for each token during inference.
The slow storage bottleneck inspired our novel pre-attention router,
which predicts required experts before attention computation,
enabling our inference engine to hide storage I/O latency through intelligent prefetching.
To further optimize storage access patterns, we introduce DP-Groups Global Load Balance Loss,
which promotes expert specialization while maintaining training stability,
allowing predictable activation patterns that enable effective expert caching strategies.
The memory limitations drove the development of our NoPE-RoPE hybrid sparse attention,
dramatically reducing KV cache requirements without sacrificing long-context understanding.

Beyond architecture, we co-design the entire inference stack for local deployment.
To fully exploit the inherent sparsity in ReGLU activation,
we develop fused sparse FFN kernels that bypass unnecessary computations in up and down projection layers.
We design a predictor for the language model head that selectively activates weight rows,
dramatically reducing computational overhead.
For memory-constrained environments, we leverage expert activation locality to strategically offload expert parameters to SSDs,
establishing an efficient I/O-computation pipeline combined with our pre-attention router that completely hides storage latency.
This holistic system design transforms theoretical sparsity into real-world speedups,
enabling \model-21B-A3B and \model-4B-A0.6B to achieve what was previously thought impossible:
over 20 tokens per second on ordinary consumer CPUs—without any GPU acceleration—using just 1 GiB and 8 GiB of memory respectively,
while achieving state-of-the-art MMLU scores that outperform even larger LLMs. Specifically, SmallThinker-21B-A3B and SmallThinker-4B-A0.6B achieves up to 86× and 19× decoding speed improvement compared to Qwen3-30B-A3B and Qwen3-1.7B, respectively, which matches the speed of in-memory baselines like Gemma3n-E4B and Gemma3n-E2B.

\model{} represents more than an engineering achievement—it's a paradigm shift in how we think about AI deployment.
By designing natively for local constraints, we demonstrate that the future of AI need not be limited by the reach of cloud infrastructure.
Instead, we can bring truly capable AI to billions of devices worldwide,
enabling a new era of private, responsive, and universally accessible artificial intelligence.
\section{Model Architecture}
\label{sec:model}

\begin{table}[t]\small
    \centering
    \caption{Summary of the {\model} architectures.}
    
    \begin{tabular}{lccccccccc}
    \toprule
            \textbf{Model} &
            \textbf{\makecell{\# of\\Layers}} &
            \textbf{\makecell{Hidden\\Dim.}} &
            \textbf{\makecell{Head\\Dim.}} &
            \textbf{\makecell{FFN\\Dim.}} &
            \textbf{\makecell{\# of Q\\Heads}} &
            \textbf{\makecell{\# of KV\\Heads}} &
            \textbf{\makecell{\# of\\Experts}} &
            \textbf{Top K} \\
    \midrule
    \model-4B-A0.6B & 32 & 1536 & 128 & 768 & 12 & 2 & 32 & 4 \\
    \model-21B-A3B & 52 & 2560 & 128 & 768 & 28 & 4 & 64 & 6 \\
    \bottomrule
    \end{tabular}

    \label{tab:nemotron-h-arch}
    \end{table}

\begin{figure}[!t]
    \centering
    \includegraphics[width=0.9\textwidth]{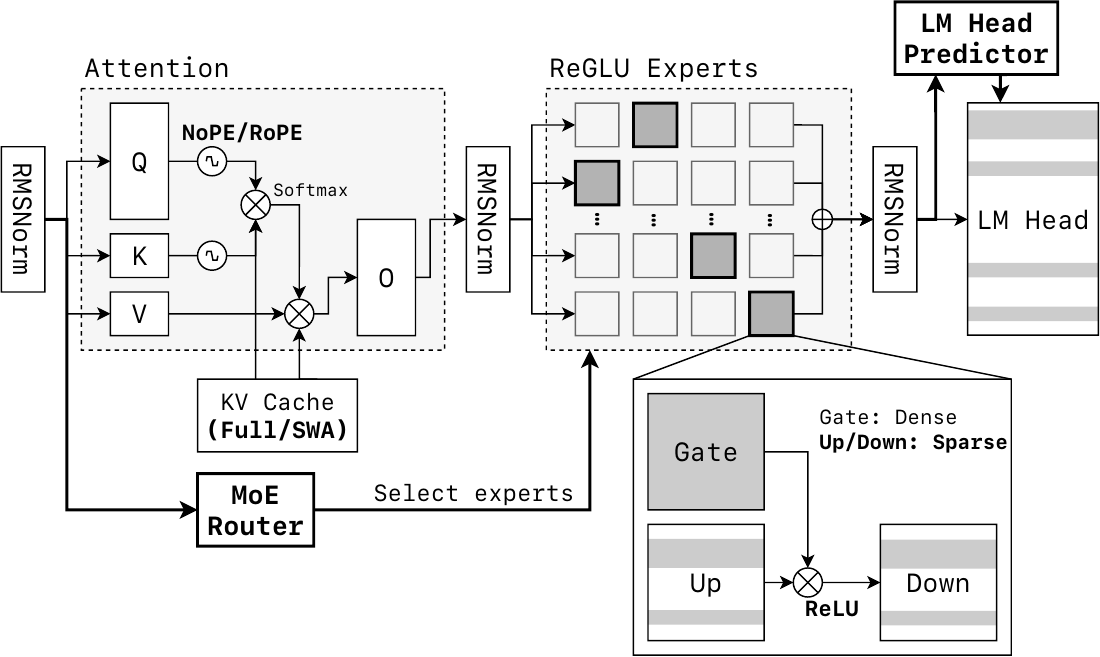}
    \caption{\model{} model architecture. Residual connections are omitted for clarity.}
    \label{fig:smallthinker-arch}
\end{figure}

{~\model} series models are based on standard decoder-only Transformer architecture with several unique variants. Table~\ref{tab:nemotron-h-arch} shows the key information on model architecture and Figure~\ref{fig:smallthinker-arch} shows the model architecture overview, where we highlight the key architectural decisions below.

\subsection{Fine-Grained Mixture of Experts}
\label{sec:FineGrainedMoE}

\paragraph{Basic Architecture of \model{}}
For MoE, prior research has demonstrated that fine-grained experts exhibit superior parameter efficiency~\citep{dai2024deepseekmoeultimateexpertspecialization}.
Accordingly, we adopt a fine-grained Mixture of Experts (MoE) architecture.
For our 4B and 21B models, we configure them with 32 and 64 experts, respectively.

\paragraph{Sparse ReGLU-Based FFN}
Although the MoE architecture already introduces structural sparsity by design,
we observe that employing activation functions from the ReLU family can induce further neuron-level sparsity within each expert~\citep{zhang2024relu, luo2024sparsing, song2024turbo}.
To further reduce computational costs and I/O, we construct each FFN expert using the ReGLU activation function.

\paragraph{Pre-Attention Router}
On-device platforms, such as AI PCs and smartphones, typically need to serve multiple applications concurrently,
which leaves a limited memory budget for LLM serving.
This constraint necessitates efficient parameter offloading.
To create a sufficient time window for prefetching the required expert parameters,
we position the MoE router module \textbf{before} the attention block.
This architectural design allows us to determine which experts are needed early,
enabling the prefetching of their weights to occur in parallel with the attention computation.

\paragraph{DP-Groups Global Load Balance Loss}
Expert specialization refers to the behavior where different experts within a model learn to process distinct types of inputs or sub-tasks. This property is highly beneficial for on-device inference because a given task will consistently activate a small, predictable subset of ``hot'' experts. This enables a hierarchical caching strategy where these active experts are stored in fast memory (DRAM), while the remaining ``cold'' experts reside on slower storage (e.g., SSD). However, the conventional load balance loss, typically applied at the micro-batch level, imposes a strong constraint that promotes uniform expert activation across the global batch. This uniformity directly counteracts the goal of specialization. To address this, we employ a more relaxed data-parallel groups (DP-Groups) Load Balance Loss. Our loss operates within smaller, partitioned DP-groups, allowing different groups to cultivate specialized experts based on their group-level global tokens. This achieves the desired functional specialization while keeping training overhead negligible.

\subsection{NoPE-RoPE Hybrid Sparse Attention}
To achieve an optimal balance between model performance and KV cache efficiency, we adopt the NoPE-RoPE Hybrid Sparse Attention architecture proposed in prior work~\citep{puvvada2025swan}.
This architecture operates on a repeating 1:3 pattern across the model's layers: one layer of global attention with No Positional Embedding (NoPE) is followed by three consecutive layers of Sliding Window Attention (SWA, window size: 4096) with Rotary Position Embedding (RoPE).
By strategically employing global attention, this hybrid design effectively curtails the KV cache footprint while largely preserving the model's long-context modeling capabilities.

\section{Pre-Training}
\model-4B-A0.6B-Base and \model-21B-A3B-Base are pre-trained on a large, primarily English corpus of high-quality data, including curated open-source content and synthetically generated data.

\subsection{Data Construction}

\paragraph{Open-Source Dataset Collection}
Following prior work such as SmolLM~\citep{allal2025smollm2}, we initiated our data construction process by collecting a diverse range of high-quality datasets from the open-source community.
For web data, we aggregated a corpus totaling 9 trillion tokens from prominent sources including FineWeb-Edu~\citep{lozhkov2024fineweb-edu}, Nemotron-CC~\citep{su2024nemotron}, 
mga-fineweb-edu~\citep{hao2025reformulation}
and the Knowledge Pile~\citep{fei2024query}.
For math datasets, we collected 1 trillion tokens, primarily from datasets such as OpenWebMath~\citep{paster2023openwebmath},
MegaMath~\citep{zhou2025megamath}, and FineMath~\citep{allal2025smollm2} and so on.
For our coding dataset, we established corpora like StackV2~\citep{lozhkov2024starcoder} and OpenCoder~\citep{Huang2024OpenCoderTO}.
After collecting these datasets, we proceeded to grade them to filter low quality corpus using a suite of corpus quality scoring models.
Considering that minor data duplication does not lead to significant degradation in model quality, we did not perform rigorous data deduplication.

\paragraph{Synthetic Data Construction}
Given the relative scarcity of high-quality,
open-source data for mathematics and code,
we employed an MGA-style methodology~\citep{hao2025reformulation} and a persona-driven methodology~\citep{ge2024persona} for data augmentation.
We simplified the overall workflow for the MGA-style synthesis.
For each domain, we first sampled a small subset of the data to generate a pool of diverse genres and audiences. After a manual curation process to select the most effective ones, we augmented each data sample in the corresponding domain by randomly pairing it with one of these pre-selected prompts. This approach streamlines the process and accelerates data synthesis.
Using these methods, we generated an additional 269 billion tokens of mathematical and code data. We are also making these synthesized code\footnote{The PowerCoding dataset: \url{https://huggingface.co/datasets/PowerInfer/PowerCoding}} and math\footnote{The PowerMath dataset: \url{https://huggingface.co/datasets/PowerInfer/PowerMath}} datasets publicly available to support further research.

\paragraph{SFT-Style Data}
Following established methodologies from models like SmolLM~\citep{allal2025smollm2} and Nemotron-H~\citep{blakeman2025nemotron},
we also incorporate SFT-style instruction-response data into the final stage of pre-training.
To build this dataset, we first collected a diverse set of open-source SFT corpora. We then augmented this collection by extracting millions of question-answer (QA) pairs directly from the highest-quality web texts identified by our scoring models during the data collection phase.

\subsection{Data Mixture}
Our pre-training corpus is organized into six high-level data categories: General Knowledge, Mathematics, Code, Web-STEM,
Chinese, and SFT-style data.

We structure our pre-training process into a three-stage curriculum. This staged approach allows us to dynamically adjust the data composition over time, progressively shifting from a broad data foundation to a more specialized and high-quality mix. In the initial stage, the model is exposed to a wide array of general data to build fundamental language and knowledge representations. As training progresses into the subsequent stages, we strategically increase the proportion of high-quality data, as identified by our scoring models, and simultaneously up-weight the mixture ratios for our specialized domains: STEM, Mathematics, and Code. For the final stage, we introduce the highest quality corpora and a rich mix of SFT-style data.

\subsection{Pretraining Hyperparameter Setting}
We trained \model-4B-A0.6B on  a token horizon of 2.5 trillion tokens and \model-21B on a token horizon of 7.2 trillion tokens. For \model-4B-A0.6B,
we used a sequence length of 4096 and global batch size of 1536 (6,291,456 tokens per batch) with a peak learning rate of 3e-4.
For \model-21B-A3B,
we used a sequence length of 4096 and global batch size of 4352 with a peak learning rate of 4.2e-4.
We used cosine learning rate decay with a minimum value equal to 10\% of the peak value, weight decay of 0.1,
and 0.9 and 0.95 for Adam $\beta_1$ and $\beta_2$, respectively. The pretrain time of 21B is only 20 days.

\subsection{Long Context Pre-Training}
{\model} employ a two-phase pre-training approach.
We began with a 4096-token context and in the final stage,
extended it to 32768 tokens for 4B models and 16384 tokens for 21B models.
This extension process, we performed concurrently with an adjustment of the Rotary Position Embedding (RoPE) base frequency from 1e5 to 1.5e6.

\section{Post-Training}

\subsection{Supervised Fine-Tuning}
Our Supervised Fine-Tuning (SFT) dataset is with serveral key areas: Knowledge-Intensive, Code, Math data.
Each category required a unique strategy for data collection and synthesis, which we break down in detail below.

\paragraph{Knowledge-Intensive Data}
To build our knowledge-intensive QA pair,
we first identified the highest-scoring portion of our general pre-training corpus using a scoring model.
With this data, we extracted a vast number of of question-answer pairs. The answers were then refined to enhance their quality,
ultimately yielding a dataset of over 10 million samples.

\paragraph{Math and Code}
Creating high-quality training data for Math and Code domains presents a unique challenge: the number of problems with verifiable ground truths is limited. To address this, we developed a multi-stage data generation pipeline focused on quality and diversity.
First, we sample core problems from a wide range of open-source datasets. To expand this seed set, we then augment these questions using a persona-based methodology~\citep{ge2024persona}. For each resulting query, we leverage Qwen3-32B (thinking) model to generate detailed, step-by-step solutions. Given that reasoning-intensive responses in these domains can be lengthy, all generated answers undergo rigorous filtering to mitigate the risk of degeneration and ensure linguistic consistency. 

\subsection{Model Merging} 
To further enhance our model's performance, we employed a model merging~\citep{goddard-etal-2024-arcees} technique inspired by previous research. After completing the SFT process, we performed linear interpolation on the weights of various checkpoints captured at different training stages. This process allowed us to find a balance between general knowledge and instruction-following fidelity.

\section{Model Evaluation}

\subsection{Model Performance}
\begin{figure}[h]
    \centering
    \includegraphics[width=0.9\linewidth]{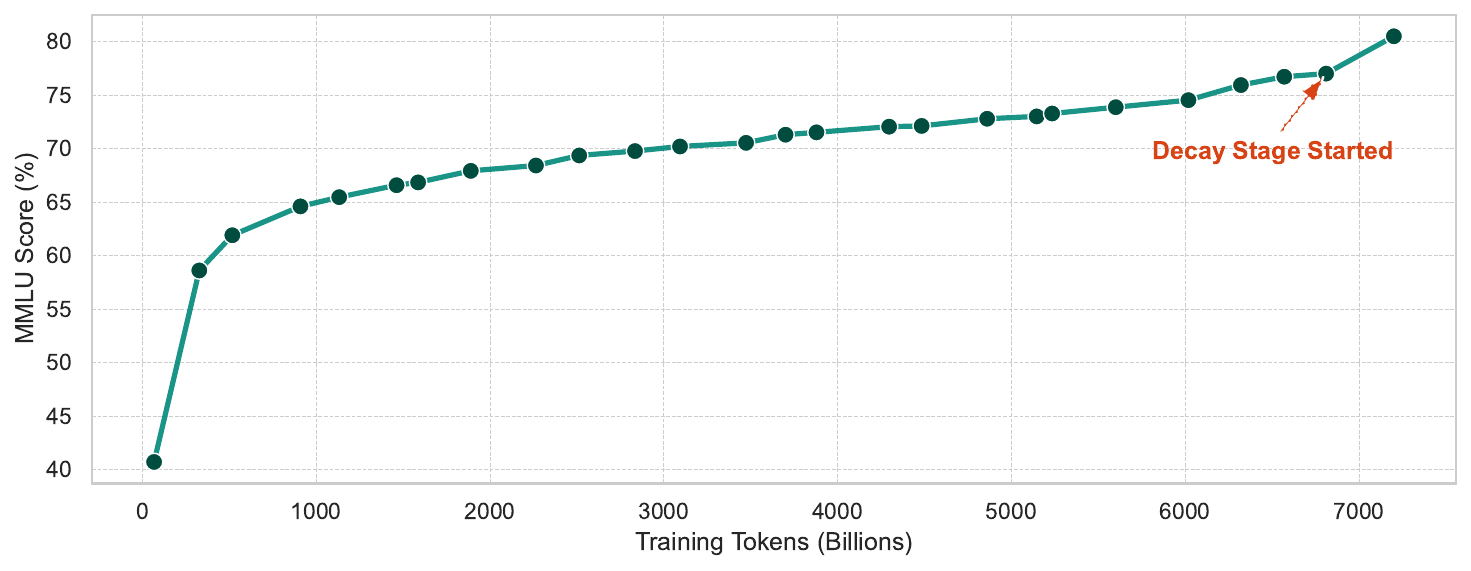}
    \caption{Learning curve of \model-21B-A3B-Base on the MMLU benchmark, showing 5-shot accuracy vs. training tokens (in billions).}
    \label{fig:mmlu-training}
\end{figure}
\paragraph{SmallThinker-21B-A3B} Our evaluation is conducted using an internal, modified version of the OpenCompass~\citep{2023opencompass}. We adopt distinct protocols for different model types: base models are evalutaed via option-based perplexity, while fine-tuned models are evaluated in a 0-shot setting using a generation-based approach where answers are extracted and then matching with ground truth. Within this evaluation setup, we compare SmallThinker-21B-A3B against baseline models including Gemma3-12B, Qwen3-8B, Qwen3-14B, and Phi4-14B, all of which activate more parameters than SmallThinker-21B-A3B. Additionally, we include the recently open-sourced MoE model Qwen3-30B-A3B as a strong baseline, which contains more total parameters than SmallThinker-21B-A3B. As shown in Table~\ref{tab:accuracy21B}, SmallThinker-21B-A3B achieves performance comparable to Qwen3-30B-A3B across general tasks (MMLU and LiveBench), mathematical tasks (GPQA-diamond and MATH-500), instruction-following tasks (IFEval), and coding tasks (HumanEval), despite having fewer total and activated parameters. Furthermore, SmallThinker-21B-A3B demonstrates significant improvements over Gemma3-12B, Qwen3-8B, and Phi4-14B, which activate more parameters during inference. The efficient parameter utilization of SmallThinker-21B-A3B enables it to finish complex tasks on resource-constrained devices. We also report the MMLU score during the pretraining phase (Figure~\ref{fig:mmlu-training}), where the model's performance exhibits a steep initial improvement, followed by a period of more gradual gains. Notably, during the final training stage corresponding to the annealing schedule, we observe a second phase of accelerated performance growth.

\begin{table}[h]
    \centering
    \caption{Comparison among SmallThinker-21B-A3B and other representative strong open-source counterparts. The highest and second-best scores are shown in \textbf{bold} and \underline{underlined}, respectively.}
    \scalefont{0.85}
    \begin{tabular}{l *{6}{c}}
    \toprule
     & \textbf{\makecell{SmallThinker\\21B-A3B}} & \textbf{\makecell{Qwen3\\14B}} & \textbf{\makecell{Qwen3\\30B-A3B}} & \textbf{\makecell{Phi4\\14B}} & \textbf{\makecell{Gemma3\\12B}} & \textbf{\makecell{Qwen3\\8B}} \\
    \midrule
    MMLU & 84.4 & 84.8 & \textbf{85.1} & \underline{84.9} & 78.5 & 80.8 \\
    GPQA-Diamond & \underline{55.1} & 50.0 & 44.4 & \textbf{55.5} & 34.9 & 38.9 \\
    MATH-500 & 82.4 & \textbf{84.6} & \underline{84.4} & 80.2 & 82.4 & 81.6 \\
    IFEval & \textbf{85.8} & \underline{85.2} & 84.3 & 63.2 & 74.7 & 83.9 \\
    LiveBench & \textbf{60.3} & \underline{59.5} & 58.8 & 42.4 & 44.5 & 49.5 \\
    HumanEval & \underline{89.6} & 88.4 & \textbf{90.2} & 87.2 & 82.9 & 85.9 \\
    \bottomrule
    \end{tabular}
    \label{tab:accuracy21B}
\end{table}

\paragraph{SmallThinker-4B-A0.6B} We compare SmallThinker-4B-A0.6B against baseline models of similar sizes, including Llama3.2-3B and Gemma3n-E4B. Additionally, we include Qwen3-0.6B, Qwen3-1.7B, Llama3.2-1B, and Gemma3n-E2B as baselines that activate more parameters than SmallThinker-4B-A0.6B. As shown in Table~\ref{tab:accuracy4B}, SmallThinker-4B-A0.6B outperforms all counterparts that activate a similar number of parameters. We also evaluate against the strong baseline Gemma3n-E4B, which activates the same number of parameters as SmallThinker-4B-A0.6B's total parameter count. Despite having significantly fewer activated parameters, SmallThinker-4B-A0.6B achieves performance comparable to Gemma3n-E4B.

\begin{table}[h]
    \centering
    \caption{Comparison among SmallThinker-4B-A0.6B and other representative strong open-source counterparts. The highest and second-best scores are shown in \textbf{bold} and \underline{underlined}, respectively.}
    \scalefont{0.85}
    \begin{tabular}{l *{7}{c}}
    \toprule
         & \textbf{\makecell{SmallThinker\\4B-A0.6B}} & \textbf{\makecell{Gemma3n\\E4B}} & \textbf{\makecell{Gemma3n\\E2B}} & \textbf{\makecell{Qwen3\\1.7B}} & \textbf{\makecell{Qwen3\\0.6B}} & \textbf{\makecell{Llama3.2\\3B}} & \textbf{\makecell{Llama3.2\\1B}} \\
    \midrule
        MMLU           & \underline{66.1} & \textbf{71.3} & 63.0 & 64.2 & 43.3 & 64.2 & 45.7 \\
        GPQA-Diamond   & \textbf{31.3} & \underline{27.8} & 20.2 & \underline{27.8} & \underline{27.8} & 24.2 & 22.7 \\
        MATH-500       & 60.6 & \textbf{69.2} & 58.6 & \underline{63.6} & 45.6 & 40.0 & 14.4 \\
        IFEval         & 69.7 & \textbf{78.4} & \underline{73.2} & 69.5 & 58.4 & 71.2 & 48.1 \\
        LiveBench      & \textbf{42.2} & 34.7 & 27.9 & \underline{35.6} & 23.1 & 15.3 & 13.5 \\
        HumanEval      & \textbf{82.3} & \underline{74.4} & 64.6 & 61.6 & 31.7 & 55.5 & 37.2 \\
    \bottomrule
    \end{tabular}
    \label{tab:accuracy4B}
\end{table}

\subsection{Expert Specialization}
\begin{figure}[h]
    \centering
    \includegraphics[width=1.0\linewidth]{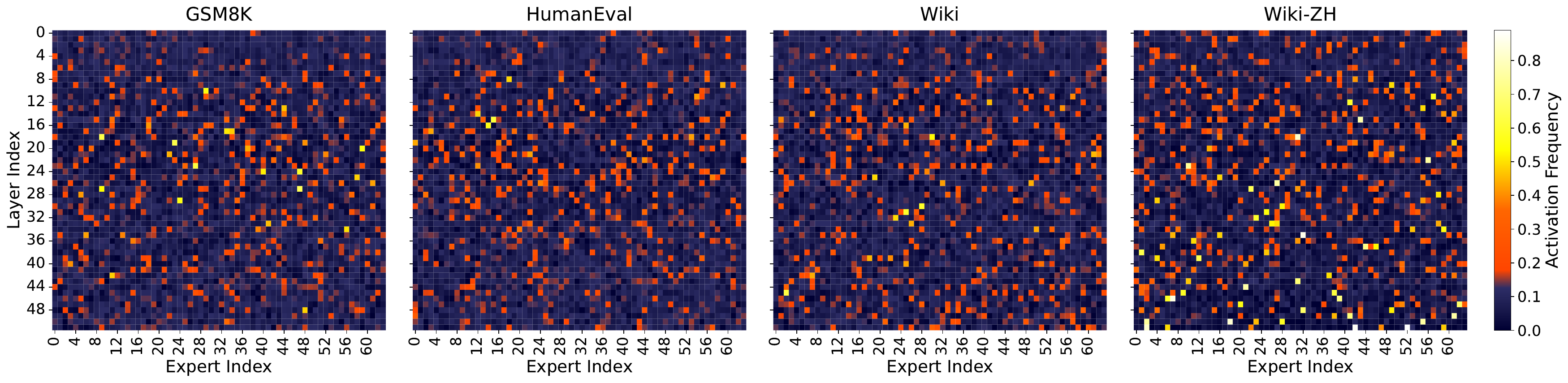}
    \caption{Expert activation frequency heatmaps of SmallThinker-21B-A3B.}
    \label{fig:21b-expert}
\end{figure}

\begin{figure}[h]
    \centering
    \includegraphics[width=1.0\linewidth]{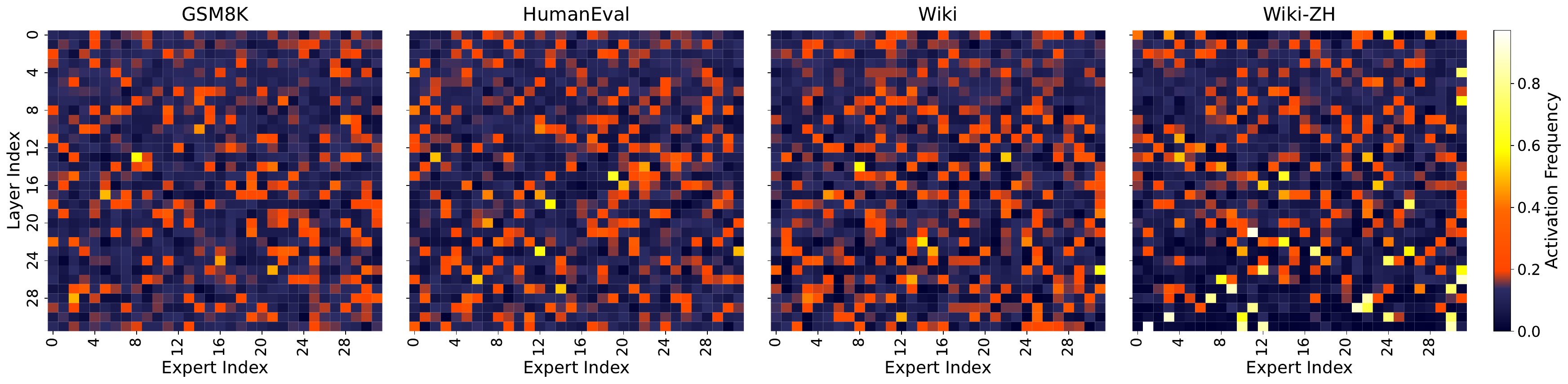}
    \caption{Expert activation frequency heatmaps of SmallThinker-4B-A0.6B.}
    \label{fig:4b-expert}
\end{figure}

We conducted comprehensive analysis of expert activation frequencies across four datasets (GSM8K, HumanEval, Wiki, Wiki-ZH) for both 4B and 21B parameter MoE variants to evaluate the specialization patterns. The activation heatmaps reveal distinct expert specialization across task domains and languages, with significant heterogeneity both across and within layers. As shown in Figure~\ref{fig:21b-expert}, The 21B model exhibits more pronounced sparsity with clearer delineation between highly active and dormant experts, indicating that increased capacity enables refined specialization. Statistical analysis confirms that 70\%-80\% of experts maintain activation frequencies below 0.14 across all datasets, while the remaining 20\%-30\% demonstrate significantly higher rates (0.4-0.6), establishing the fundamental sparse activation nature of MoE architectures. The similar characteristics are also suggested for the 4B model in Figure \ref{fig:4b-expert}. These predictable activation patterns, characterized by consistent hotspots and dataset-specific behaviors, provide a robust foundation for implementing expert caching strategies and task-adaptive prefetching mechanisms to optimize inference performance.

\subsection{Neuron-Level Sparsity}
\label{sec:neuron-level-sparsity}
\begin{figure}[h]
    \centering
    \subfloat[][\centering{\model-4B-A0.6B}]{
        \includegraphics[width=0.36\linewidth]{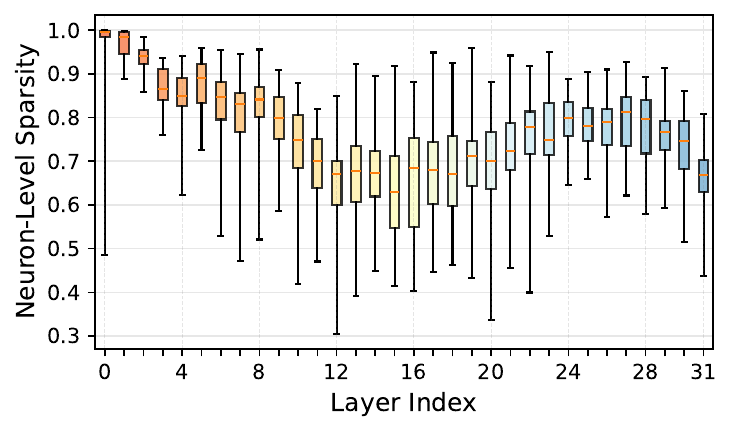}
        \label{fig:neuron-sparsity-4b}
    }
    \hfil
    \subfloat[\centering{\model-21B-A3B}]{
        \includegraphics[width=0.58\linewidth]{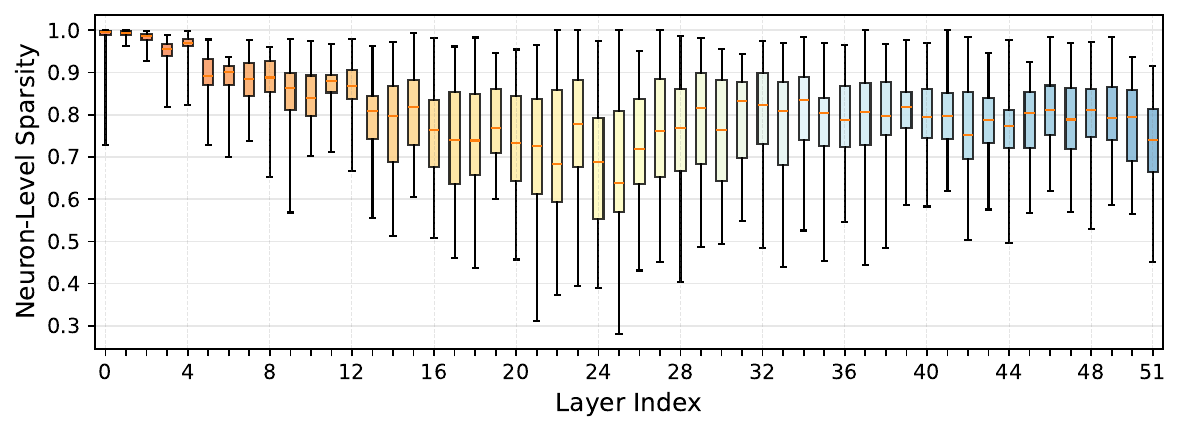}
        \label{fig:neuron-sparsity-21b}
    }
    \caption{The neuron-level sparsity across layers for SmallThinker model family. Each box plot represents the distribution of neuron sparsity values within experts at a given layer index. The boxes show the interquartile range (IQR) from the first quartile (Q1) to the third quartile (Q3), with the orange line indicating the median. Whiskers extend to the minimum and maximum values in the data. The color gradient from orange to blue represents the progression from early to later layers.}
    \label{fig:neuron_sparsity_comparison}
\end{figure}

We analyzed intra-expert neuron-level sparsity of both \model-4B-A0.6B and \model-21B-A3B. As shown in Figure~\ref{fig:neuron_sparsity_comparison}, both models maintain remarkably high sparsity across all layers, with median values consistently above 0.6. Initial layers exhibit near-perfect sparsity (0.9-1.0), and most neurons within deeper-layer experts exhibit sparsity levels exceeding 0.6. The phenomenon confirms that even if the expert is routed, over 60\% of neurons remain inactive. This pervasive neuron-level sparsity, induced by the ReLU activation function can be utilized further the accelerate the decoding speed.
\section{Inference Framework for Local Devices}

\subsection{Memory-Efficient Inference}

The \model{} is specifically designed for devices with limited memory capacity, such as personal computers, smartphones, and low-end embedded devices. Its expert-level sparsity allows part of the model parameters to be offloaded to secondary storage (e.g., SSD). 
As offloading inevitably incurs non-negligible overhead from I/O operations, we utilize pre-attention routers to initiate I/O requests early and overlap SSD accesses with attention computations.

\paragraph{Expert Offloading}
To fit into restricted memory on local deployment environments (e.g., PC, smartphones and embedded ARM processors), we implement a parameter offloading mechanism that offloading a portion of expert parameters to SSD storage. The primary challenges arise from the substantial throughput disparity between DRAM-SSD communication bandwidth and inference performance requirements.
To address this, we leverage the locality of expert activation by DP-group global load balance loss (\S\ref{sec:FineGrainedMoE}) to implement a shared expert cache across all MoE layers. With LRU replacement policy, we identify and retain the most frequently accessed experts during inference, which efficiently reduces potential SSD access bottlenecks.

\paragraph{Expert Prefetching Pipeline}
To further address the substantial latency overhead from DRAM-to-storage I/O during LLM inference, we leverage expert activation information from pre-attention routing to proactively prefetch expert weights during attention computation. This enables overlapping of I/O operations with computational tasks, maximizing I/O bandwidth utilization and hiding memory access latency. During MoE FFN computation, we set up a pipeline that interleaves expert loading with computation. By parallelizing I/O tasks and computation tasks, it efficiently minimize SSD access impact on inference throughput. Combined with optimized caching policies that achieve high cache hit rates, these techniques efficiently prevent I/O latency from blocking inference, significantly improving performance on resource-constrained devices.

\subsection{Sparse Inference}
To address the limited computational capacity of local deployment environments and enhance inference efficiency, \model{} comprehensively leverages the inherent sparsity characteristics of contemporary LLMs. Our analysis reveals that ReLU-activated Gated Linear Units (ReGLU) and Language Model Head (LM head) computations exhibit substantial sparsity, thereby providing significant opportunities for accelerating inference in resource-constrained local deployment scenarios.

\paragraph{ReGLU Sparsity}
As discussed in \S~\ref{sec:neuron-level-sparsity}, through the integration of ReGLU, \model{} attains an additional 60\% sparsity enhancement beyond the inherent expert sparsity provided by the MoE architecture.
During the inference, we employ a selective computation strategy wherein all entries of the gate matrices are evaluated, while neurons in the up and down projection matrices are computed discriminately based on the ReLU activation function outputs. 
To actualize substantial performance gains, we have developed highly optimized fused sparse ReGLU FFN kernels that leverage SIMD (Single Instruction, Multiple Data) vectorization instructions for efficient parallel processing.

\paragraph{LM Head Sparsity}
LM head computations constitute a substantial proportion of total inference latency and exhibit computational complexity of $O(HV)$, where $H$ represents the hidden size and $V$ denotes the vocabulary size. This computational requirement necessitates floating-point operations (FLOPs) scaling beyond $10^8$, presenting a significant performance bottleneck. To mitigate the LM head computation during inference, we introduce a dedicated predictor module specifically designed to identify vocabulary rows with high activation probability during inference. 
Based on the predictor's output probabilities, the inference system selectively computes logits corresponding to activated rows while setting the remaining logits to zero, thereby eliminating unnecessary computational overhead and optimizing resource utilization. 

\subsection{Inference Performance Evaluation}
To validate the effectiveness of our proposed techniques, we implement all optimizations within the PowerInfer\footnote{\url{https://github.com/SJTU-IPADS/PowerInfer/tree/main/smallthinker}} framework~\citep{song2024powerinfer, xue2024powerinfer}, which provides a robust foundation for sparse inference and expert caching mechanisms.
The inherent characteristics of \model{} facilitate local deployment by optimizing both memory utilization and inference latency,
thereby enabling efficient model execution on resource-constrained devices with limited memory and computational capabilities.

\paragraph{End-to-End Performance} 

We evaluate the inference throughput of \model{} and other models using PowerInfer framework with Q4\_0 quantization. As demonstrated in Table~\ref{tab:decoding-21b}, the in-memory inference performance of \model{} achieves up to 30.19 tokens/s on PC. On smartphones and embedded ARM processors, the inference throughput exhibit improvement by up to 65\% compared to Gemma3n-E4B. This performance enhancement is attributed to the comprehensive utilization of expert-level sparsity in conjunction with the inherent sparsity characteristics of ReGLU feed-forward networks and the language model head, which collectively minimize computational overhead and optimize resource efficiency.

\begin{table}[h]
    \footnotesize
    \centering
    \caption{Decoding performance (in tokens/s) of \model-21B-A3B, Qwen3-30B-A3B, and Gemma3n-E4B. All models are quantized with Q4\_0.}
    
    \begin{tabular}{l|ccc}
        \toprule
            \textbf{Device} & \textbf{\model-21B-A3B} & \textbf{Qwen3-30B-A3B} & \textbf{Gemma3n-E4B} \\
        \midrule
        PC (i9 14900K) & 30.19 & 33.52 & 21.93 \\
        OnePlus 13 (Snapdragon 8 Gen 4) & 23.03 & 20.18 & 16.58 \\
        RK3588 (A55$\times$4+A76$\times$4) & 10.84 & 9.07 & 7.37 \\
        Raspberry Pi 5 (A76$\times$4) & 6.61 & OOM & 4.01 \\
        \bottomrule
    \end{tabular}
    
    \label{tab:decoding-21b}
\end{table}

\begin{table}[h]
    \small
    \centering

    \caption{Decoding performance (in tokens/s) of \model-4B-A0.6B, Qwen3-0.6B, Qwen3-1.7B, and Gemma3n-E2B across various platforms. All models are quantized with Q4\_0.}

    \scalebox{0.9}{
        \begin{tabular}{l|cccc}
        \toprule
            \textbf{Device} & \textbf{\model-4B-A0.6B} & \textbf{Qwen3-0.6B} & \textbf{Qwen3-1.7B} & \textbf{Gemma3n-E2B} \\
        \midrule
        PC (i9 14900K) & 108.17 & 148.56 & 62.24 & 36.88 \\
        OnePlus 13 (Snapdragon 8 Gen 4) & 78.99 & 94.91 & 41.00 & 27.06 \\
        RK3588 (A55$\times$4+A76$\times$4) & 39.76 & 45.93 & 20.29 & 12.50 \\
        Raspberry Pi 5 (A76$\times$4) & 28.77 & 27.44 & 11.08 & 6.66 \\
        RK3576 (A53$\times$4+A72$\times$4) & 15.10 & 15.29 & 6.09 & 3.80 \\
        RDK X5 (A55$\times$8) & 7.23 & 13.32 & 6.35 & 3.46 \\
        RK3566 (A55$\times$4) & 6.33 & 9.76 & 4.15 & 2.45 \\
        \bottomrule
        \end{tabular}
    }
    \label{tab:decoding-transposed}
\end{table}

    


\paragraph{Expert Offloading Performance}

For memory-constrained devices where available memory cannot accommodate the complete model weights, we conduct comprehensive evaluations with 8 GiB and 1 GiB memory limit for \model-21B-A3B model and \model-4B-A0.6B model respectively. As shown in Table~\ref{tab:offloading-decoding-21b}, \model-21B-A3B achieves up to 85$\times$ higher throughput than Qwen3-30B-A3B, matching the inference speed of Gemma3n-E4B which fully resident in memory. This performance advantage stems from the expert prefetching pipeline, which effectively hides SSD access latency through parallel attention computation, thereby maintaining inference latency comparable to in-memory inference scenarios. In contrast, alternative models experience significant latency penalties due to memory swapping overhead and inefficient random fine-grained read operations on SSD storage, which are inherently suboptimal for the sequential access patterns typical in neural network inference workloads. Table~\ref{tab:offloading-decoding-4b} suggests the similar improvement of \model-4B-A0.6B. 

\begin{table}[!h]
    \small
    \centering
    \caption{Decoding performance (in tokens/s) of \model-21B-A3B, Qwen3-30B-A3B, and Gemma3n-E4B, with 8 GiB memory limit. All models are quantized with Q4\_0.}
    
    \begin{tabular}{l|ccc}
    \toprule
        \textbf{Device} & \textbf{\model-21B-A3B} & \textbf{Qwen3-30B-A3B} & \textbf{Gemma3n-E4B} \\
    \midrule
    PC (i9 14900K) & 20.30 & 10.11 & 21.93 \\
    OnePlus 13 (Snapdragon 8 Gen 4) & 15.50 & 0.18 & 16.58 \\
    RK3588 (A55$\times$4+A76$\times$4) & 8.56 & 6.32 & 7.37 \\
    \bottomrule
    \end{tabular}
    
    \label{tab:offloading-decoding-21b}
\end{table}

\begin{table}[!h]
    \small
    \centering
    \caption{Decoding performance (in tokens/s) of \model-4B-A0.6B, Qwen3-1.7B, and Gemma3n-E2B, with 1 GiB memory limit. All models are quantized with Q4\_0.}
    
    \begin{tabular}{l|ccc}
    \toprule
        \textbf{Device} &
        \textbf{\model-4B-A0.6B} &
        \textbf{Qwen3-1.7B} &
        \textbf{Gemma3n-E2B} \\
    \midrule
    PC (i9 14900K) & 29.99 & 2.66 & 36.88 \\
    OnePlus 13 (Snapdragon 8 Gen 4) & 20.91 & 1.09 & 27.06 \\
    RK3588 (A55$\times$4+A76$\times$4) & 15.04 & 1.00 & 12.50 \\
    \bottomrule
    \end{tabular}
    
    \label{tab:offloading-decoding-4b}
\end{table}

    


    
    

\section{Limitations and Future Work}
\label{sec:conclusion}

One primary limitation of \model is the scale of its pretraining data. Our training corpus is considerably smaller than those used for state-of-the-art models. Consequently, our models may exhibit limitations in its knowledge breadth and performance on a wider array of tasks.
Addressing this is a key priority for our future work. We plan to scale our pretraining efforts by expanding the training dataset to enhance the model's general-purpose capabilities and ensuring its competitiveness in a rapidly advancing field.

Furthermore, \model{} has only undergone supervised fine-tuning and has not been further aligned using Reinforcement Learning from Human Feedback (RLHF). This means the model's responses, while instruction-following, may lack the nuanced quality, helpfulness, and safety guarantees that RLHF can provide. As a key part of our future work, we plan to implement an RLHF pipeline to further refine and polish the quality of the model's responses, ensuring they are better aligned with user preferences.

\section{Authors}

Yixin Song, Zhenliang Xue, Dongliang Wei, Feiyang Chen, Jianxiang Gao, Junchen Liu,\\
Hangyu Liang, Guangshuo Qin, Chengrong Tian, Bo Wen, Longyu Zhao, Xinrui Zheng,\\
Zeyu Mi (\url{zeyumi@zenergize.ai}), Haibo Chen (\url{haibochen@sjtu.edu.cn}).

\bibliography{ms}

\end{CJK*}
\end{document}